# Deep Aramaic: Towards a Synthetic Data Paradigm Enabling Machine Learning in Epigraphy

Andrei C. Aioanei,[1] Regine Hunziker-Rodewald,[1] Konstantin Klein,[2] Dominik L. Michels[3]

[1] University of Strasbourg (France)

[2] University of Amsterdam (Netherlands)

[3] KAUST Visual Computing Center (KSA)

**Abstract**

Epigraphy increasingly turns to modern artificial intelligence (AI) technologies such as machine learning (ML) for extracting insights from ancient inscriptions. However, scarce labeled data for training ML algorithms severely limits current techniques, especially for ancient scripts like Old Aramaic. Our research pioneers an innovative methodology for generating synthetic training data tailored to Old Aramaic letters. Our pipeline synthesizes photo-realistic Aramaic letter datasets, incorporating textural features, lighting, damage, and augmentations to mimic real-world inscription diversity. Despite minimal real examples, we engineer a dataset of 250,000 training and 25,000 validation images covering the 22 letter classes in the Aramaic alphabet. This comprehensive corpus provides a robust volume of data for training a residual neural network (ResNet) to classify highly degraded Aramaic letters. The ResNet model demonstrates 95% accuracy in classifying real images from the 8th century BCE Hadad statue inscription (KAI 214). Additional experiments validate performance on varying materials and styles, proving effective generalization. Our results validate the model's capabilities in handling diverse real-world scenarios, proving the viability of our synthetic data approach and avoiding the dependence on scarce training data that has constrained epigraphic analysis. Our innovative framework elevates interpretation accuracy on damaged inscriptions, thus enhancing knowledge extraction from these historical resources.

## 1. Introduction

Deciphering ancient scripts through the study of inscriptions, known as epigraphy, provides valuable insights into historical languages and cultures. However, the automated recognition and interpretation of ancient writing systems pose considerable challenges for ML techniques [ASP19, SAPS+23]. In particular, Old Aramaic scripts present a highly complex task for algorithmic analysis.

Aramaic has been used as a script for over three millennia, evolving into distinct regional styles [Nöl01]. The Aramaic alphabet, first used in the 9th century BCE, has 22 letter-signs. The Aramaic alphabet has evolved to include localized variations in writing conventions [Tei15]. Surviving Aramaic inscriptions display tremendous diversity based on temporal, geographic and linguistic factors [Fit95]. For instance, ancient Imperial Aramaic was used for official inscriptions during the Achaemenid era in the 1st millennium BCE [Gze15]. Later forms like Jewish Babylonian Aramaic and Jewish Palestinian Aramaic emerged following the exile of Jewish populations [Sok19]. This evolution resulted in scripts of late dialects like Syriac, Mandaic, and Samaritan in the Common Era [MP98].

Besides linguistic differences, Aramaic texts were inscribed on various materials, such as stone, ceramic, papyrus, parchment, and metal, each imparting a distinct visual quality [Bor09]. For example, Stone inscriptions were either incised or in relief. Inscriptions on papyrus were typically displayed in dark ink on a yellowish background, on parchment in dark ink on a white background, and on ceramic in dark or red ink on a brown background. Environmental factors over centuries of exposure have also eroded many inscriptions.

Some promising applications of deep neural networks have recently emerged for analyzing ancient scripts like Old Greek, Cuneiform, and Egyptian hieroglyphs [ASP19; BM22; BCCF+22]. However, a persistent barrier is the lack of sufficient training examples, with datasets often limited to 10,000s of images. Significant amounts of annotated letters are challenging to access due to various factors, including the preservation condition of ancient inscriptions, limited accessibility, and the time-consuming nature of manual annotation. Synthetic data generation provides a potential solution by producing labeled corpora for teaching ML models [KWPP+23; Tsi22]. Recent advances in procedural modeling and computer graphics enable the creation of simulated environments with high visual realism [KWPP+23]. However, frameworks tailored to the complexities of historical scripts like Old

Aramaic remain lacking. Manual annotation is extremely challenging, given the temporal and geographic breadth of the ancient scripts. This scarcity of labeled training data has constrained computational approaches to recognizing Aramaic scripts.

This paper presents a novel pipeline for the synthetic generation of annotated Old Aramaic letter datasets covering various styles and materials. The synthetic data is used to train deep neural networks for classifying Aramaic glyphs[1] within inscriptions by algorithmically simulating factors such as linguistic variations, material differences, and erosion effects. Our approach bridges the interdisciplinary gap between ML and cultural heritage, providing a scalable solution for deciphering ancient inscriptions where real-world training data is scarce.

For our analysis, we will test on several Old Aramaic inscriptions across the Northern Levant region [SS11]. For instance, the mid-8th-century monumental Hadad statue from Gerçin preserves an inscription in 34 lines (KAI 214). The inscription has been intentionally damaged and is, in addition, in certain places, considerably worn, making many of its letters hard to discern, even with modern high-resolution visual techniques.

This paper is organized as follows. Section 2 reviews the related work in the field, highlighting the current gaps our work aims to address. Section 3 discusses the challenges in Old Aramaic letter recognition. Section 4 presents our methodological approach, from synthetic data generation to model training and testing. This is followed by section 5, which details our experiments and results. Lastly, sections 6 and 7 provide discussion and conclusions by summarizing our findings and their implications for epigraphy.

## 2. Related Works

The application of ML algorithms to analyze ancient scripts has seen growing interest in recent years. However, a persistent challenge hampering the performance of modern deep learning models on these tasks is the scarcity of large-scale, representative training datasets. We review relevant literature on ML for ancient languages, existing datasets, and synthetic data generation.

---

[1] In this paper, we use the distinction between symbols, glyphs, and signs to describe the different levels of ancient writing systems precisely emulated through our synthetic data generation approach. A symbol refers to the conceptual written mark or character that makes up the vocabulary of a script. It is the abstract unit of communication. A glyph is a rendered graphical variant or visualization of a particular symbol. It covers the artistic realization of the symbol. A sign denotes a functional unit of meaning within a script's vocabulary. It represents the mapping between symbols and their linguistic meanings. For example, the conceptual modern mark 'A' is a symbol representing a particular sound. But this symbol may be drawn or carved in different artistic glyph forms. As a unit within the writing system, 'A' also serves as a phonetic sign, mapping that symbol to the /a/ phoneme through practice and convention [Ferr20, COG18].

## 2.1 Machine Learning for Ancient Scripts

The application of machine learning to analyze ancient scripts has seen growing interest, but progress has been constrained by limited training data. Ancient script datasets typically comprise 10,000s of examples, far smaller than the volumes needed to train modern deep neural networks effectively. This scarcity poses obstacles to the performance of modern deep learning models, which typically require far larger volumes of data.

Early works focused on small subsets of ancient writing systems. Edan [Eda13] classified Cuneiform signs using k-Nearest Neighbors on 1,500 examples. Mostofi and Khashman [MA14] experimented with shallow neural networks on 5,000 cases. Can et al. [COG16] worked with 10,000 Maya glyphs. These initial studies demonstrated potential but lacked diversity.

Subsequent research expanded in scale and scope. Firmani et al. [FMMN18] used deep segmentation and classification methods for 25,000 examples of Latin texts. Franken and van Gemert [FvG13] proposed an image retrieval approach for Egyptian hieroglyphs classification. Bogacz and Mara [BG20] augmented limited Cuneiform data to improve classification across historical periods. These pioneering efforts demonstrated potential but were restricted in scope.

More recent research has expanded to larger datasets, which enabled more complex models. Swindall et al. [SCHK+21] collected over 490,000 annotated ancient Greek letters and evaluated convolutional networks. The best results came from fine-tuning a Residual Network pretrained on real images. Haliassos et al. [HBSQ+20] detected Egyptian hieroglyphs in papyri using deep CNNs. Rizk et al. [RRRK22] designed a capsule network architecture for classifying Phoenician letters. Moustafa et al. [MHHA+22] built an end-to-end hieroglyph translation system based on computer vision.

However, data scarcity remains a fundamental challenge. Less resourced ancient languages have often been compensated through data augmentation or generative models. Nguyen et al. (2021) denoised damaged Cham inscriptions using encoder-decoder networks. Bogacz and Mara [BM22] used augmented data to classify Cuneiform across historical periods. Swindall et al. [SPKW+22] improved Greek text recognition through augmented data. However, these approaches do not fully capture the linguistic and physical complexities of ancient writing systems.

Overall, the growing availability of digitized data resources has enabled deep neural networks to push state-of-the-art analytics for ancient scripts. However, small, imbalanced datasets with limited diversity remain pressing challenges, motivating innovative solutions to generate representative training data where manual annotation is infeasible. Our work addresses this by proposing a tailored paradigm to generate synthetic Aramaic data where manual annotation is infeasible.

### 2.2 Training Datasets for Ancient Scripts

Applying modern ML to ancient scripts requires massive labeled datasets, often millions of examples, to effectively train models. However, compiling such large annotated corpora is extremely labor-intensive, resulting in scarce training data resources for most ancient languages.

Currently, documented datasets for ancient writing systems are limited, with examples numbering in the 10,000s, insufficient for cutting-edge algorithms. For instance, pioneering character recognition research on Ancient Greek relied on just 10,000 annotations spanning centuries of texts [ASP19; ASSB+22]. Studies of Egyptian hieroglyphs used 17,000 computer-generated images rather than true inscriptions [BCCF+22; BCFL+21; GPFC+23]. Cuneiform classification experiments were conducted with only 1000 glyphs [BM22; LSYH+21]. Efforts have been made to compile Aramaic script resources. However, the documented corpora cover only limited linguistic and temporal diversity subsets. Many are limited to less than 50,000 examples [FSSF+15, RRRK22, and SGFS+20]. These datasets also suffer from metadata inconsistencies, class imbalance, unrealistic data, and inadequate diversity. The examples frequently focus on a narrow subset of the script's temporal and geographic span.

Some efforts have expanded resources for specific ancient languages. The PROIEL offers 187,000 syntactic annotations for Ancient Greek [SAPS+23], while the Index Thomisticus Treebank contains over 60 million words from Medieval Latin [SAPS+23]. However, diversity issues persist, and most ancient languages remain under-resourced. Documented Aramaic corpora cover limited periods with inconsistently labeled examples.

In summary, while ancient script datasets are growing, their scarcity, imbalance, lack of diversity, and inconsistent metadata continue to impede computational analysis. Our work addresses this by algorithmically generating comprehensive, annotated Old Aramaic data tailored to train ML models unconstrained by real-world limitations.

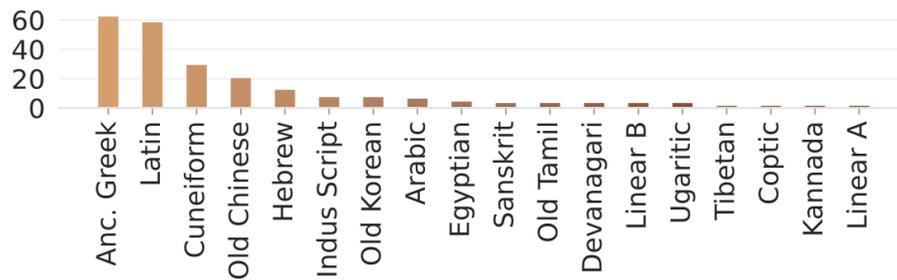

Figure 1: Distribution of publications by ancient language, showing an imbalanced distribution attributable to differences in data availability. Languages with fewer than 2 publications are excluded [SAPS[+]23].

### 2.3 Synthetic Data Generation

Synthetic data offers a promising solution for generating abundant labeled datasets to train ML models where sufficient real-world data is difficult to obtain. It involves algorithmically creating simulated data that closely approximates the characteristics of real examples [Arc18; Nik21]. Virtual simulation of environments and objects enables the creation of large-scale corpora with automatic ground truth annotation while controlling diversity, rare events, and other factors [Tsi22].

In contexts like epigraphy, where real-world training data is limited, synthetic data can help overcome these constraints by providing unlimited volumes of labeled data to train models.

#### 2.3.1 Benefits of Synthetic Data

Synthetic data provides valuable advantages over reliance solely on real-world examples [Arc18; Tsi22; KWPP+23]:

- Automatic Ground Truth Generation: Synthetic data enables automatic annotation of class labels, segmentation masks, bounding boxes, etc. This is crucial for tasks requiring expensive manual labeling.
- Controlled Diversity: Simulated data can cover the target domain densely by generating various scenarios and conditions. This improves model robustness.
- Scalability: Synthetic data is highly scalable, allowing large datasets where content complexity and size can be varied easily. This helps address data scarcity.

- Mitigating Data Scarcity: Synthetic data generation is a practical solution when real-world data collection is constrained by logistics, costs, or effort. It provides abundantly available training resources.
- Rare Events Simulation: Uncommon edge cases or challenging scenarios can be intentionally synthesized to improve model resilience.
- Privacy: Synthetic data avoids privacy issues associated with using personal information in real datasets.
- Balancing Datasets: Imbalanced class distributions can be addressed by intentionally generating more examples for underrepresented categories.
- Cost-Efficiency: Synthetic data creation is typically far more cost-effective than exhaustive manual data annotation or curation.
- Control over Data Characteristics: Researchers can precisely define synthetic data parameters and characteristics tailored to their problem.

According to Gartner - a leading technology research and advisory company that provides data-driven insights and predictions across industries - by 2024, 60% of data used in these projects will be synthetically generated. By 2030, synthetic data is predicted to completely overshadow real data in AI models. Gartner anticipates this growth due to the increasing need for large training datasets as ML adoption spreads across sectors. As real-world data collection remains expensive and time-consuming, yet ML models demand ever-growing data volumes, synthetic data provides a scalable solution. As synthetic generation methods improve, models trained on artificial datasets are expected to perform on par or better than those relying solely on real-world data. In fields like epigraphy, where real-world training data is limited, these predictions highlight the vital role synthetic data could play in unlocking the potential of ML. Our research exemplifies this in the context of Old Aramaic inscriptions.

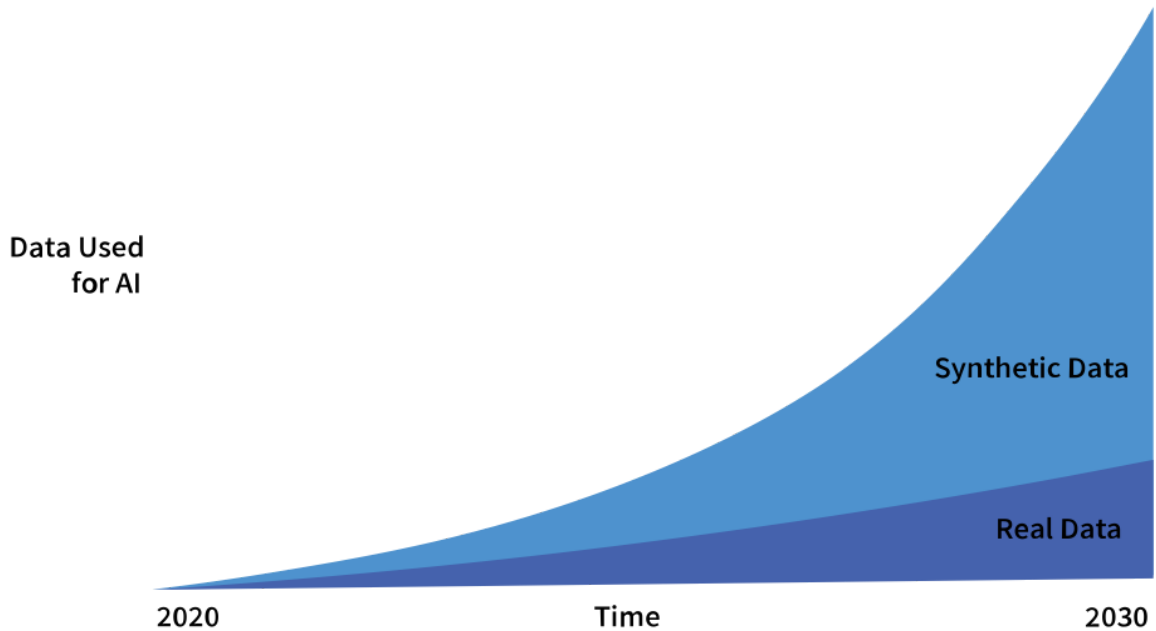

Figure 2: The estimated growth in the use of synthetic data for AI in line with Gartner's prediction.

### 2.3.2 Synthetic Data for Computer Vision

In computer vision, synthetic datasets have been used to train object detectors for urban driving by generating photorealistic street scenes with simulated cars, pedestrians, buildings, and traffic signs. It controls imaging conditions and camera parameters, which is challenging with real-world data [JBMN+17; TPAB+18; TTSX+18].

### 2.3.3 Synthetic Data for Text Recognition

For training text recognition networks, synthetic data provides more flexibility in incorporating font, style, and lexical variances compared to scanned documents [JSVZ14; GVZ16]. Approaches like generative adversarial networks have been used to mimic handwriting styles and distort synthetic text to simulate real-world noise [CBHC19]. Vögtlin et al. proposed a procedural generation method for mimicking degradations in medieval manuscript images (VDPA+21). But this centered more on stains and holes than material and linguistic properties. Character structure, stroke connectivity, rhythm, slant, and proportion are difficult to simulate. In other words, accurately encapsulating the nuances of human writing remains an open challenge.

### 2.3.4 Synthetic Data for Ancient Scripts

While synthetic data generation has been applied in diverse domains, its use in analyzing ancient scripts remains limited. Most efforts focus narrowly on textual features without encapsulating the physical attributes of inscriptions.

Assael et al. [ASP19; ASSB+22] synthesized damaged Greek texts to train models for restoring missing characters. However, this did not account for writing medium factors. Barucci et al. [BCFL+21, see Figure 3.1] generated augmented Egyptian hieroglyph datasets by transforming real examples. But this fails to mimic the material support and degradation effects over time. Lazar et al. [LSYH+21] modeled Cuneiform glyphs parametrically but did not incorporate degradation effects. Vögtlin et al. [VDPA+21] simulated stains and holes in medieval manuscripts without considering material-specific aging. Focusing only on typographic features fails to encapsulate damage and material wear. Rizk et al. [RRRK21, Figurine 3.2] introduced a framework using synthetic data generation and augmentation to train a neural network for classifying Phoenician script letters. However, their pipeline relied on creating synthetic letters as white glyphs on black backgrounds rather than real-world images. While demonstrating the potential of synthetic data when real examples are scarce, their simulated images may not fully capture the complex factors affecting real inscriptions, such as damages and surface noise. Using simplified synthetic inputs risks training models that do not generalize well to real-world Phoenician inscriptions and texts.

Some recent studies attempt to compensate for scarce training data through synthesis. Papavassileiou et al. [PKO23] artificially generated incomplete Linear B sequences to boost their generative infilling model. However, this approach does not fully capture the visual intricacies of ancient writing systems.

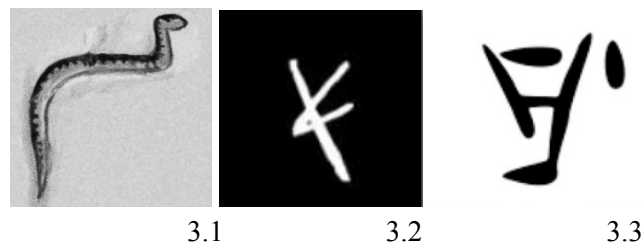

            3.1            3.2            3.3

Figure 3: (1) example of an image from Egyptian hieroglyph dataset (BCFL[+]21); (2) example of an image from an Old Aramaic dataset (RRRK21); (3) example of an image from a Cypro-Minoan dataset (CTVF22).

In summary, current synthetic data techniques for ancient scripts focus narrowly on textual properties. However, accurately modeling the complex factors that physically transform inscriptions over centuries remains an open challenge. Our work addresses this gap by algorithmically generating damaged Aramaic letters exhibiting diverse materials and erosion effects.

Strategically tailoring synthetic data generation to the nuances of ancient documents can unlock the potential of machine learning where real-world resources are limited. Extending our approach to other scripts and materials could provide training data covering the breadth needed for real-world script analysis. However, achieving sufficient realism to encapsulate physical degradation and diversity requires domain expertise and interdisciplinary collaboration. Our work demonstrates the value of this synergy in making leading machine learning techniques viable for ancient script decipherment where manual annotation is infeasible.

## 3. Challenges of the Aramaic Script

The Aramaic script presents unique challenges for developing automated recognition systems due to its extensive palaeographic, geographic, and material evolution over centuries. We discuss three key dimensions along which the script diversified into multiple styles that could pose difficulties for computational analysis.

### 3.1 Paleogeographic Evolution

The ancient Aramaic script underwent significant transformations across different eras due to changing writing tools, inscription substrates, and socio-cultural influences [Gze15, Sta20]. For instance, early 9th century BCE Old Aramaic inscriptions from sites like Tel Dan and Zincirli already display considerable angularity and stroke variation between regions [Gib71, Nav87, LS13]. The 8th-century BCE inscription from the Hadad statue exhibits idiosyncratic ligatures and conjoined letter forms reflecting a localized palaeographic style adapted to the inscription's basalt stone surface [Nav86]. Imperial Aramaic of the Achaemenid Persian era developed its own elaborate conventions suited for monumental display and bureaucratic use [Gze15]. Further diversification followed in the regional cursive scripts of the Common Era, shaped by the new writing materials of ink and parchment [Gze15].

This remarkable temporal evolution presents challenges for machine learning analysis. Training data must encapsulate style variances without bias toward one period's prevalent forms. For example, the 'lamed' symbol had as many as five handwritten variants during the Imperial Aramaic period alone [Gib71, Beyer 1986]. Models aiming to recognize this visual diversity need broad representations in their training data. Failure to capture such chronological transformations risks reduced accuracy on unfamiliar styles.

### 3.2 Geographic Diversity

In addition to temporal variations, synchronic analysis reveals pronounced geographic diversity in the Aramaic script [Nav87]. Aramaic letters demonstrate noticeable differences in stroke proportions, angularity, and ornamentation between regional scribal traditions [Gib71, Nav86]. For instance, the angular 'qoph' in 9th century BCE Aramaic texts from Tel Dan contrasts sharply with the oval 'qoph' found at Fort Shalmaneser [Bir18]. This geographic variance extended beyond major dialectal divisions to city-level writing styles [Nav87]. The Hadad statue's intricate ligatures and conjoined letters on the rough basalt surface reflect localized conventions in Sam'al [Nav86]. Models lacking diverse training data risk spatial biases that impede recognition of inscriptions outside the training distribution.

### 3.3 Material Variances

Aramaic inscriptions are preserved on diverse materials like stone, ceramic, papyrus, and metal, each imparting distinct visual properties [Gib71; Gze15]. For instance, the Hadad statue's tough basalt medium allowed intricate carvings but also caused conjoined letters from shallow stroke depths. The statue's coarse granular surface also imparts visual noise absent on smoother media. The ink spreads and fades faster on porous papyrus than on impermeable stone tablets. Training data must incorporate such medium-specific traits to ensure generalisability. Simulating diverse materials is, therefore, crucial.

### 3.4 Deterioration Effects

Historical events deliberately targeted inscriptions, destroying sections with letters on statues [Naʾ86]. Centuries of exposure to environmental factors have deteriorated many inscriptions. For instance, windborne abrasive particles can gradually abrade stone surfaces, leading to erosion of stroke edges and loss of letter contours [DR69]. Over time, environmental elements have eroded many inscriptions, causing some letters to fade or become indistinguishable from their background. Soiling, staining, and accretion deposits create visual noise. Such complex aging effects require careful modeling for training ML systems.

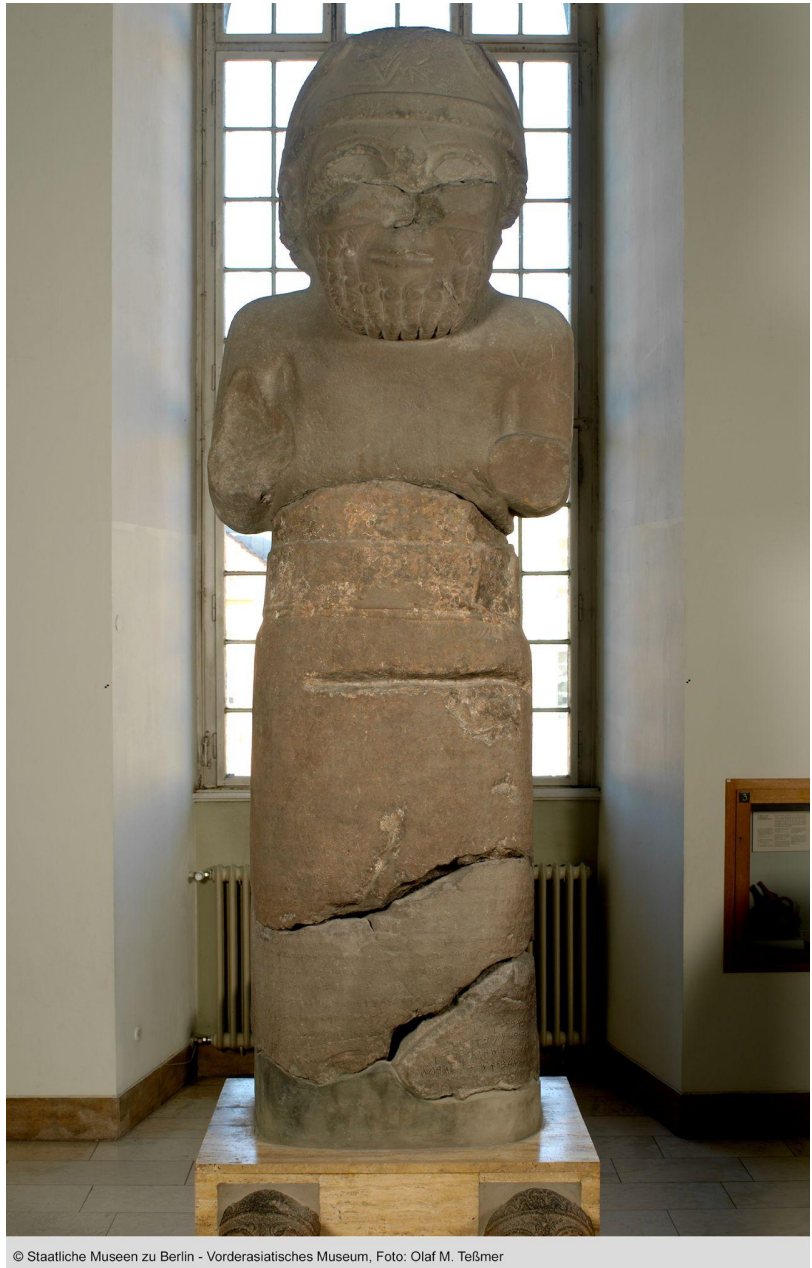

Figure 4: The Hadad statue (© Staatliche Museen zu Berlin - Vorderasiatisches Museum. Foto: Olaf M. Teßmer).

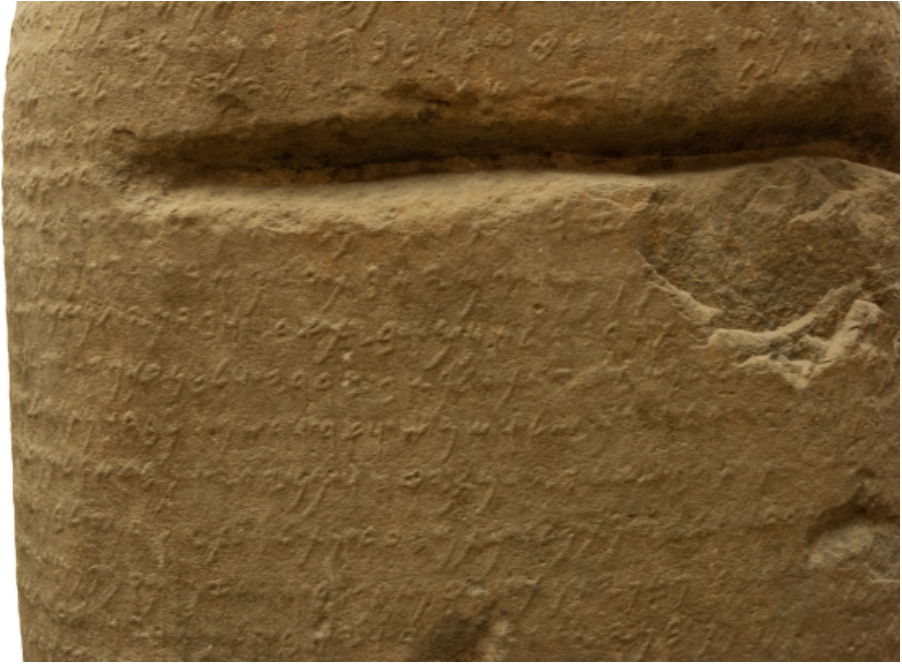

Figure 5: A section of the Panamuwa inscription on the Hadad statue (© Staatliche Museen zu Berlin - Vorderasiatisches Museum. Foto: Peter Fornaro).

In summary, when the synthetic data does not capture the complexity and variability of real-world data, a model might fail to generalize to real-world data. This is called a 'domain gap' between synthetic and real data. This refers to the differences in characteristics inherent to the capturing process, like noise or sensor-specific parameters, between computer-generated images and their real counterparts. Bridging this domain gap is a significant challenge, as it requires a deep understanding of both the generation process and the specific requirements and limitations of the machine learning model. This is particularly problematic in epigraphy, where ancient inscriptions exhibit immense diversity in script styles, materials, damage, and other variables. Solutions like our tailored synthetic approach can help overcome these limitations by algorithmically generating synthetic variability critical for training ML models.

## 4. Methodology

This section describes the methodological approach adopted. It explains the synthetic data generation pipeline with varying textures, lighting conditions, simulated damage, dataset preparation, network architecture choice, and model training and testing procedures. This method allows the creation of labeled datasets of sufficient size and diversity to train high-capacity deep neural networks despite the scarcity of real-world examples. Following this approach, we design a robust deep learning model to classify and segment Aramaic letters.

Our framework comprises the following key stages:

1. Procedural generation of 3D letter models incorporating different styles derived from epigraphic analysis.

2. Photorealistic rendering of the letter models using graphics software, with controlled randomization of parameters like textures and lighting.

3. Introduction of artificial aging effects like erosion, staining, and surface noise to mimic real-world deterioration.

4. Data augmentation techniques to further increase the variability of the synthetic dataset.

5. Final training dataset containing diverse labeled example images for each Imperial Aramaic letter class, capturing the complexities outlined earlier.

6. Implement and optimize a deep convolutional neural network architecture for classifying Old Aramaic letters from photographs.

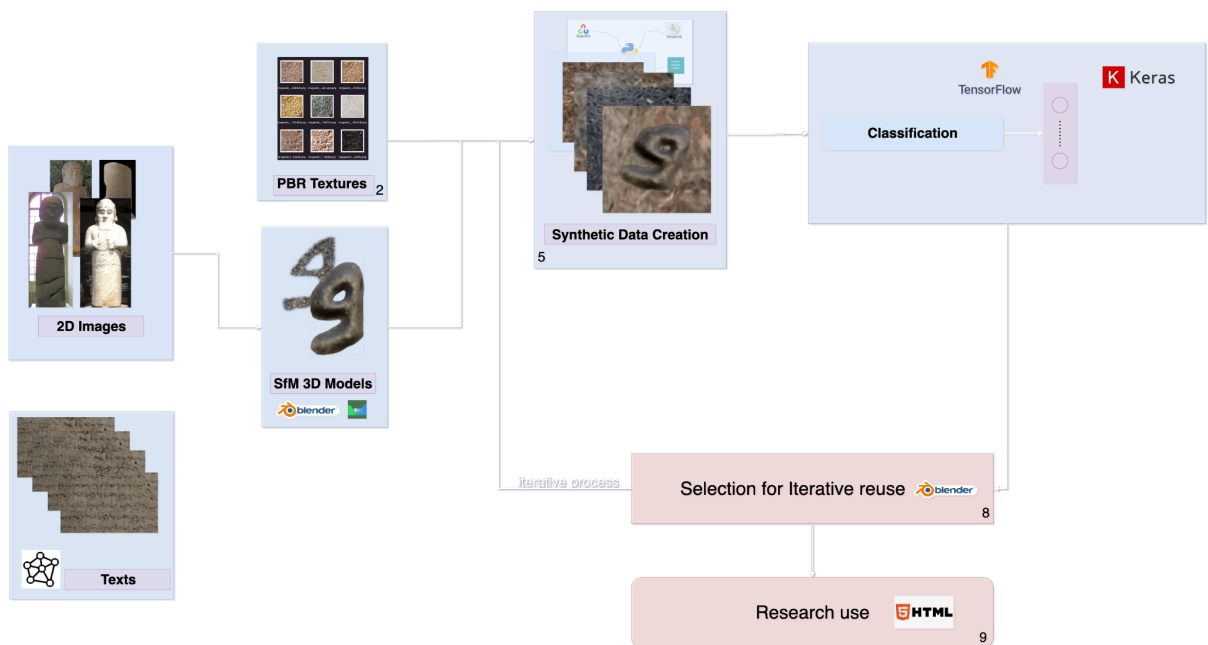

Figure 6: Diagram outlining the workflow stages.

### 4.1 Synthetic Data Generation Pipeline

Our synthetic Aramaic letter generation pipeline involves the following steps:

### 4.1.1 Letter Modeling

The first stage is the procedural modeling of 2D and 3D letter models representing the diverse styles and forms of Old Aramaic letters across different inscriptions. This is achieved by:

- Collecting examples of each Aramaic letter from inscriptions spanning significant periods like Old Aramaic, Imperial Aramaic, etc.

- Cleaning and preprocessing the collected letters to create consistent isolates of each symbol. This involves segmentation, skew/distortion correction, and noise removal.

- Performing image analysis on the segmented letters to identify common topological features and class-specific shape characteristics. We extract the visual archetypes particular to each letterform.

- Vectorizing the cleaned letters by robustly fitting smooth contours to the segmented binary shapes. The vectors encode key properties like strokes, junctions, loops, and topology.

- Further abstracting these concrete letter vectors into parametric outlines encoding higher-level structure. We define control points, widths, and style variables to synthesize new instances procedurally.

- Implementing simple geometric extrusion of the 2D outlines to create 3D mesh models with depth and relief to enable realistic rendering while preserving the Vector DNA.

- Quantitatively analyze variances in letter proportions and styles across different corpora to derive modeling style parameters. We capture critical stylistic dimensions like aspect ratio, stroke contrast, etc.

The result is a comprehensive set of Aramaic letter models covering the breadth of handwritten styles and regional conventions evident in surviving inscriptions. This robust modeling pipeline balances abstraction with retaining the uniqueness of ancient letter forms.

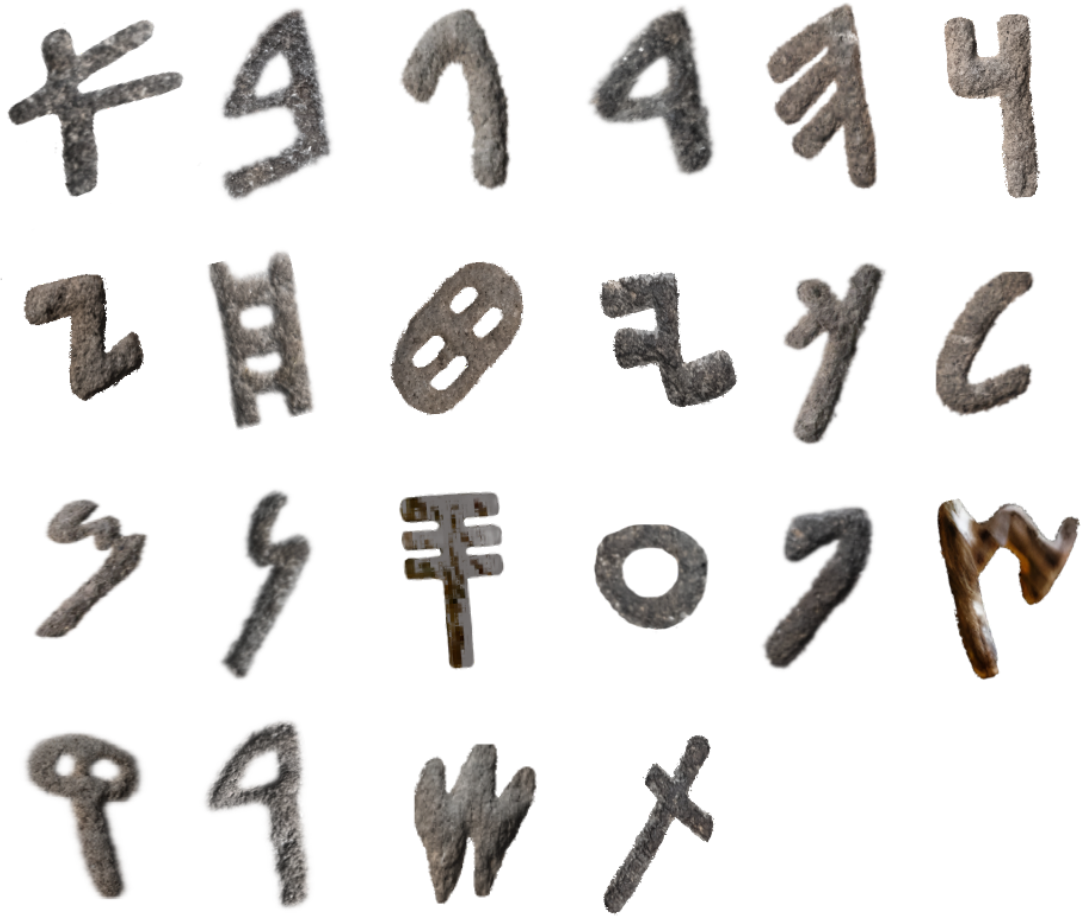

Figure 7: The letter models for the Old Aramaic script.

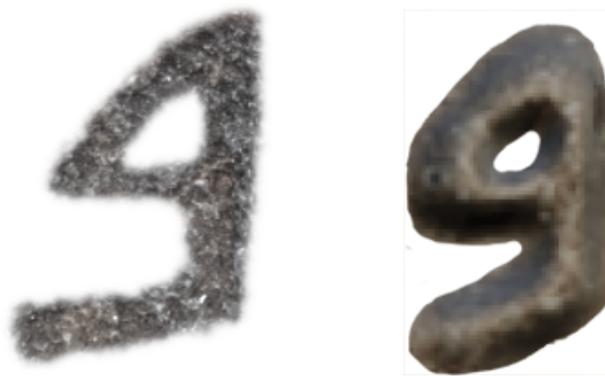

Figure 8: The bet letter model of the Old Aramaic script.

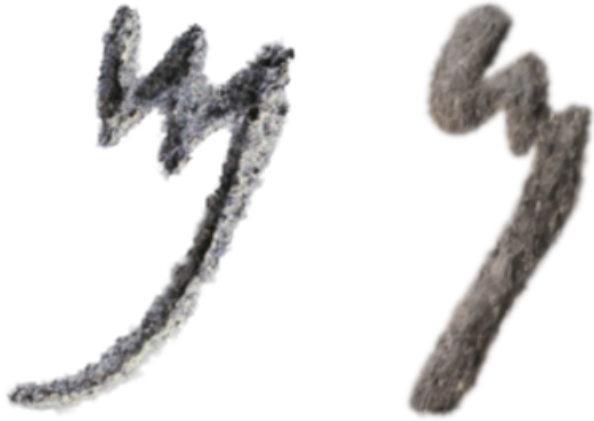

Figure 9: The mem letter model from the Kerak and the Hadad statue inscription.

### 4.1.2 Appearance Simulation

The letter models are then rendered photo-realistically using graphics software by randomizing various parameters:

- **Surface Textures**: Different procedural materials like stone, papyrus, and parchment are simulated and mapped onto the letter 3D meshes. We simulated the background also by starting with a blank image of 256 × 256 pixels and filling it with color patterns that closely resembled the materials of real-world materials such as basalt.

- **Scene Lighting**: Directional and ambient illumination conditions are randomized by configuring light source types, intensities, and colors to mimic real-world scenarios.

- **Camera Viewpoints**: The virtual camera position relative to the letter model varies across renders to capture different perspective distortions.

- **Post-processing Effects**: Camera effects like depth of field, lens distortions, and motion blur are algorithmically applied to increase realism. A Gaussian filter was subsequently applied to blur these elements. The 2D Gaussian function is defined as:

$$G(x,y) = \frac{1}{2\pi\sigma^2} \exp\left(-\frac{x^2 + y^2}{2\sigma^2}\right)$$

where x and y are the pixel location, and σ is the standard deviation of the Gaussian distribution. We then drew several circles of random sizes at random positions on the image to create letters of varying sizes.

Together, these enhance visual diversity across renders while maintaining a photorealistic style reminiscent of true Aramaic inscriptions.

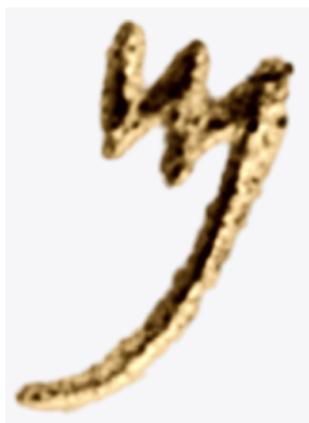

Figure 10: Applying lighting variations to the letter mem of Figure 9.

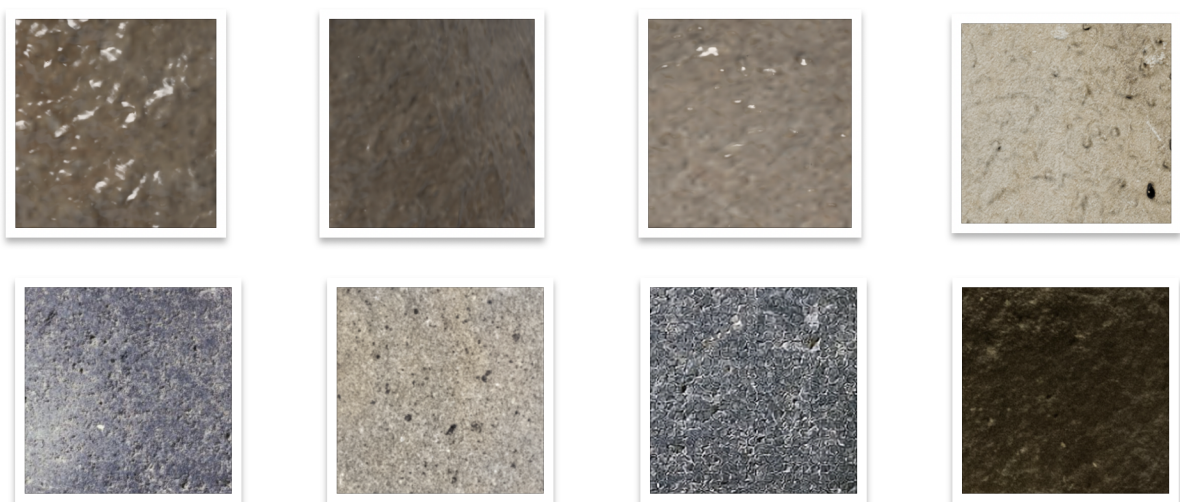

Figure 11: Synthetic backgrounds.

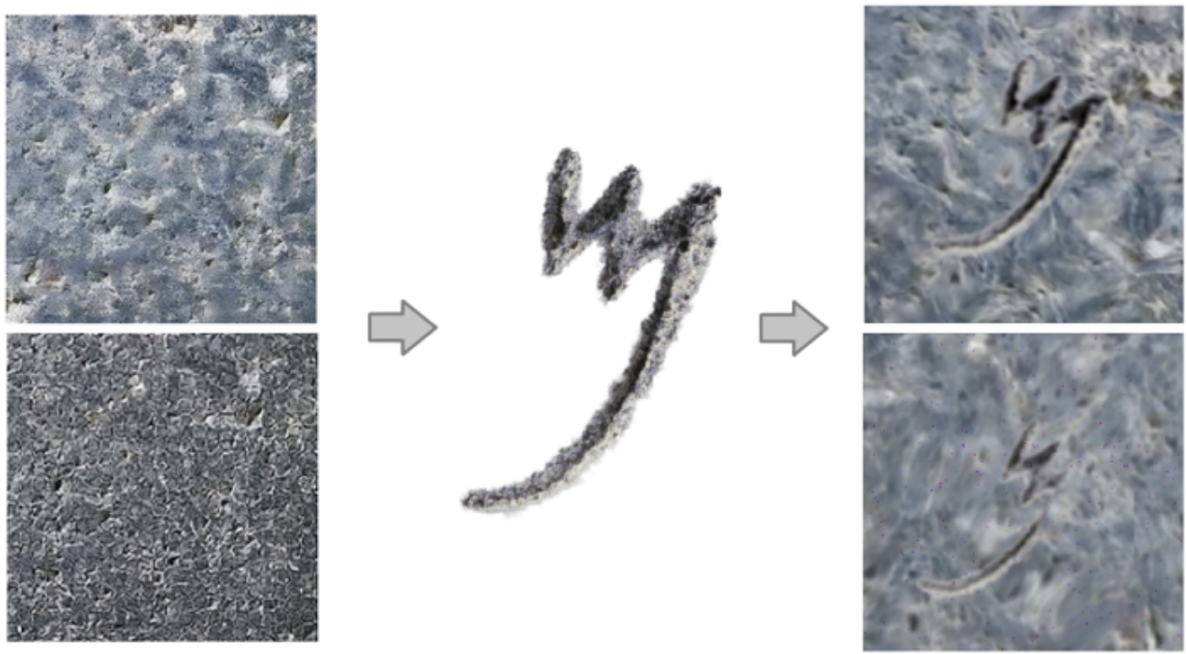

Figure 12: The synthetic data generation workflow developed to generate synthetic images of Old Aramaic letters. Different operations are included in the process, such as changes in contrast, brightness, and hue, as well as the transformations of horizontal flip, random crop, and Gaussian blur.

### 4.1.3 Damage Simulation

The rendered letter models are artificially aged by applying simulated deterioration effects:

- **Erosion**: Surface erosion is mimicked by gradually abrading the 3D mesh to create smoothed edges and topological noise.

- **Staining/Accretion**: Realistic stains, mineral deposits, and color variations are synthesized using procedural texturing.

- **Fading**: Loss of pigmentation over time is simulated by reducing texture contrast and blending letters with the base surface.

- **Fracturing/Chipping**: Small portions of the 3D geometry are procedurally removed to mimic missing stone.

- **Noise**: Camera noise, film grain, and focus effects are algorithmically added to mimic optical artifacts.

This comprehensive damage simulation allows the creation of realistic models of Old Aramaic inscriptions in various states of preservation.

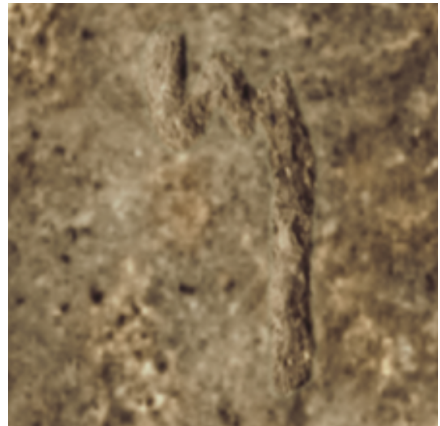

Figure 13: A mem letter simulated as eroded by gradually abrading the 3D mesh to create smoothed edges and topological noise, blending the letter with the base surface.

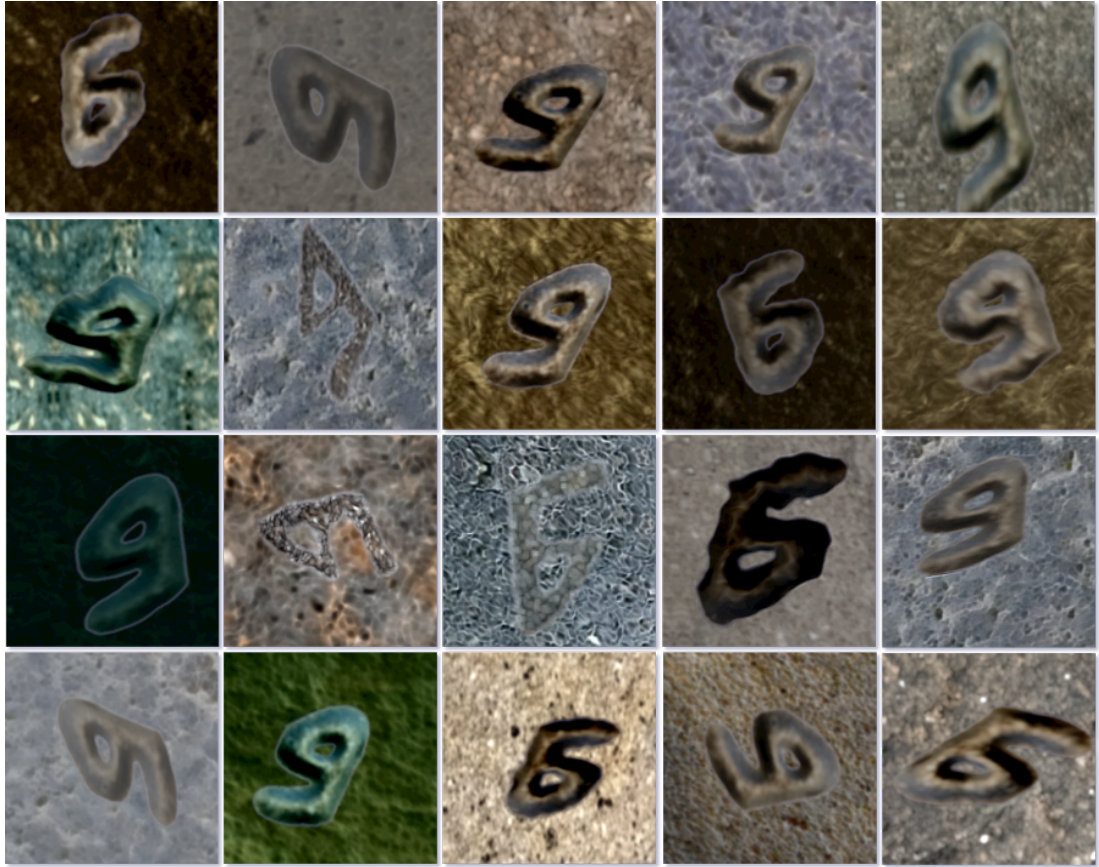

Figure 14: Samples of bet letters from a synthetic dataset.

### 4.1.4 Data Augmentation

When training a neural network on synthetic data for application on real data, a challenge often arises known as the "domain gap" [TPAB+18]. This discrepancy occurs due to differences between the data distributions of actual letter photographs and synthetic letter model renderings. It is primarily attributable to the nuances in modeling and rendering artifacts, as the synthetic rendering of letters cannot entirely match the photographic quality of real letters. To bridge this domain gap, we employed data augmentation techniques. The rendered letter images are augmented using common data expansion techniques:

- **Affine Transforms**: Images are randomly rotated, skewed, and projected into different perspectives. Additionally, we applied a random degree of Gaussian blur, flipped the image horizontally at random, and cropped it using a randomly positioned bounding box with a minimum size of 256 × 256 pixels. We also performed random geometric transformations (e.g., mirroring, rotating, zooming).

- **Color Adjustments**: Hue, saturation, brightness, and contrast are varied across images.

- **Noise Injection**: Realistic noise like speckles, blotches, and Gaussian blur is synthetically added.
- **Edge Deformations**: Letters are warped and distorted to expand shape variability. The image passed through a degradation pipeline to form the final synthetic data. Furthermore, we randomly degraded the data to enhance regularization, accounting for image quality and type variations. This degradation pipeline was adapted from the works of Ingle et al. (2019), a scalable handwritten text recognition system, and Weir et al. [WTWC21], an automated recognition system of hand-drawn hydrocarbon structures using deep learning, which both leveraged large datasets of online data for offline handwritten text recognition by applying aggressive degradation.

We intentionally made our augmentation and degradation more aggressive than one would typically find in real-world images to span the maximum dataset subspace, aiming to make the distribution as broad as possible. As previously described, the synthetic data generation pipeline stages were designed to map the synthetic distribution onto the distribution of real-world hand-drawn Aramaic letter structures.

Extensive augmentation is performed to generate a maximally diverse training dataset from the base procedural models. While augmentation does increase diversity, it can sometimes reduce the network's performance on the training dataset (KWPP+23). However, this trade-off is acceptable because it simultaneously improves the performance of new, unseen images. More advanced augmentation methods employ neural networks to transform an input image into a completely new one (KWPP+23; AJGV21).

### 4.2 Network Architecture Description

Our project aims to transform the input (image representations of Aramaic letters) into the desired output (their classifications). We employ a neural network model that encodes each input image into a high-dimensional vector, processes it, and then transforms it into a low-dimensional output vector, which is decoded into the corresponding classification.

The vector space of encoded input images is denoted as 'I', with individual input images represented as 'i $\in$ I'. We define the subspace of real input images as 'R $\subset$ I' and the subset of synthetically generated input images as 'S $\subset$ I'. 'L' represents the space of all possible labels, e.g., 'L = {alep, bet, …}'. This classification process is illustrated in Figure 14.

The training dataset 'D' comprises image-label pairs, where each image 'i' has an associated ground truth label 'G(i)'. This mapping, 'G: I → L', applies to both real and synthetic images. However, the network 'Nw', which is parameterized by its weights 'w', maps images to predictions 'p'. A prediction is correct when 'p' equals 'G(i)'.

It is important to note that the goal is not to find the correct mapping 'J → L' for images in the dataset 'D', but rather a general mapping 'I → L' for all possible input images. This process is known as generalization. A network that does not generalize well is said to overfit the training dataset. Overfitting often results from a lack of diversity in the training data. In previous section 'Data augmentation', we offered a strategy to mitigate overfitting.

In our research, we employed a custom architecture that was implemented using the TensorFlow and Keras library and built on top of the pretrained ResNet152V2 model for image classification. The model's architecture and training procedure are detailed below:

The ResNet152V2 model was initialized with pre-trained weights from the ImageNet dataset [DDSL+09]. The input size for the model was set to 224x224 pixels with three channels to represent RGB color images. The original top layers of the ResNet model were not included, as custom layers were added to tailor the model for our specific classification task.

The model's architecture was adjusted to freeze the base model's weights, making them untrainable during our process. We selectively unfroze the top 30 layers of the base model, except the Batch Normalization layers, and marked them as trainable, allowing these layers to fine-tune during the training process.

Following the base model, a separable convolutional layer with a kernel size of 3x3, 'relu' activation function, and padding set to 'same' was added. This layer was chosen for its ability to handle spatial and depth-wise features separately, which can be advantageous for computational efficiency and model performance.

The output from this layer was then passed through two different pooling layers: a Global Average Pooling and a Global Max Pooling layer. These pooled outputs were then concatenated and subjected to two successive stages of Batch Normalization and Dropout with a rate of 0.5 to mitigate overfitting.

The final dense layer was configured to output a vector of length equal to the number of classes in our dataset (22), utilizing the softmax activation function for multi-class probabilistic outputs. The model's outputs were solely based on this classifier, though it is designed to allow additional output layers if needed for other tasks.

We used the Adam optimizer with an Exponential Decay learning rate schedule for the optimization process. This allowed for a high initial learning rate that decayed over time to fine-tune the model parameters. We also introduced additional metrics (Precision, Recall, and F1-Score) beyond accuracy to evaluate the model's performance. We used the categorical cross-entropy loss function to measure training loss, and the Adam optimizer for network training.

To prevent overfitting and optimize the model's performance, we introduced Early Stopping, based on validation accuracy, and Model Checkpointing, saving the model with the best validation accuracy during the training process. The model was trained for 30 epochs, with performance evaluated against a separate validation set.

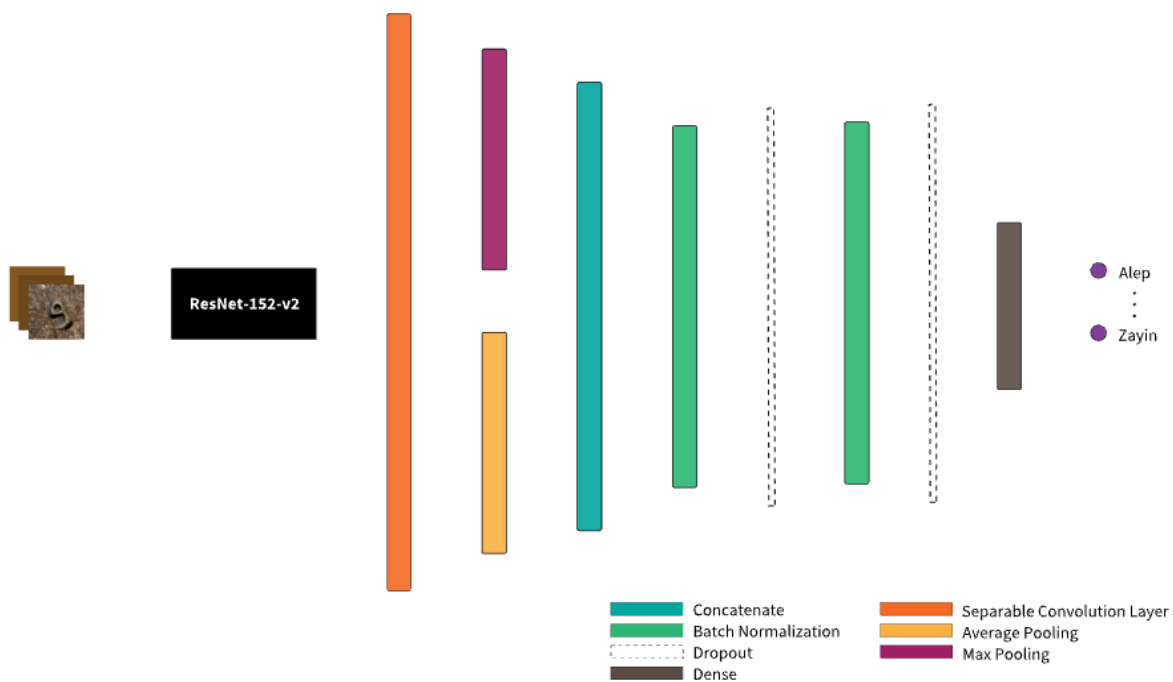

Figure 15: The architecture of our ML classification process. The process initiates with an input image, which is then converted into a vector. This vector is subsequently passed through the input layer of our pre-trained ResNet152V2 model. The resulting output vector is decoded, which maps the outputs to probabilities for each of the 22 classes.

A separable convolutional layer operates differently than a standard convolutional layer [CH23]. Instead of performing a full 3D convolution, a separable convolution first applies a single 2D convolution for each input channel and then uses a pointwise convolution (1x1) to combine the outputs. This results in fewer computations, making it computationally more efficient, and it reduces overfitting due to having fewer parameters. The output from the separable convolutional layer was then pooled and passed through further stages of the model as described in the previous discussion. This feature of our architecture improves the overall efficiency and effectiveness of the model, thus playing a vital role in our research findings.

The learning rate, beta parameters, and number of neurons have been meticulously selected to optimize the model's learning ability and efficiency. The number of classes represents the diverse categories into which the images can be classified. The pooling option 'avg' for the base model indicates average pooling in the final pooling layer before the fully connected layer in ResNet152V2. All of these constant variables are vital parameters that influence the model's learning and its performance on the image classification task.

### 4.4 Training the Model

The model was trained using a variety of dataset sizes to determine the volume of data needed to achieve the desired recognition accuracy. Datasets of sizes ranging from 150,000 to 500,000 images were used, with the data split between training and validation sets. The results of this proof-of-concept training are depicted in Fig. 15, which illustrates the positive correlation between dataset size and recognition accuracy. Furthermore, an analysis of the model's performance on individual images aided in understanding what additional features need to be incorporated into the synthetic images.

In the training process, we utilized the categorical cross-entropy loss function and the Adam optimizer, chosen due to their effectiveness in classification tasks and capability to handle large datasets, respectively. To prevent overfitting, we employed strategies such as data augmentation and early stopping, in addition to using a validation set.

We discovered that a dataset comprising 250,000 labeled images could achieve an out-of-sample (test set) accuracy of over 92%. When the dataset size was increased to 500,000 images, the model's accuracy peaked at 98%. These results confirm that the chosen network architectures can learn Aramaic letters from synthetically generated images.

The goal during training is to identify the optimal weights that correctly map input images to their labels. The training process is influenced by hyperparameters, 'h,' which include the learning rate, the batch size for the stochastic gradient descent optimizer, and data augmentation parameters.

Our network architecture consists of several linear and non-linear layers, each performing a set of mathematical operations parameterized by the network's weights. The network used in our case has 2,130,514 weights, determined through numerical optimization during the network training process.

### 4.5 Testing Methodology

The model was rigorously tested for accuracy and reliability. For this purpose, we utilized a distinct dataset composed of real images of letters, completely separate from the training and validation datasets. This testing dataset comprised around 30 images per class, providing a balanced representation of all the classes.

We used accuracy as our primary performance metric, as it directly reflects the proportion of images correctly classified by the model. It is worth noting that other metrics, such as precision, recall, or F1-score, could also be used depending on the specific requirements of the task.

In the testing phase, the model yielded 92% accuracy. However, we conducted multiple rounds of testing using different splits of the dataset, a method known as cross-validation.

The final model was chosen based on its high performance across all testing scenarios. This involved taking into account not just the average accuracy but also the consistency of performance to ensure our the model was robust to different data distributions.

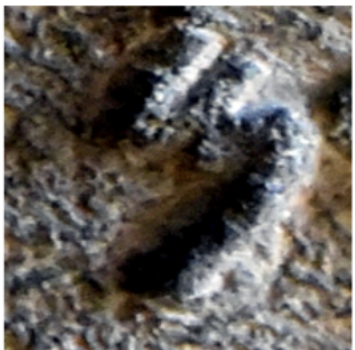
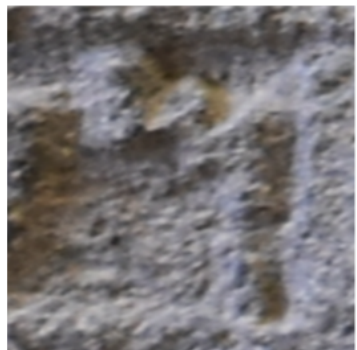
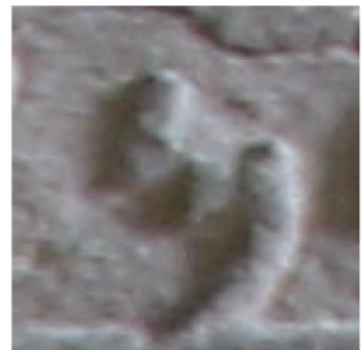
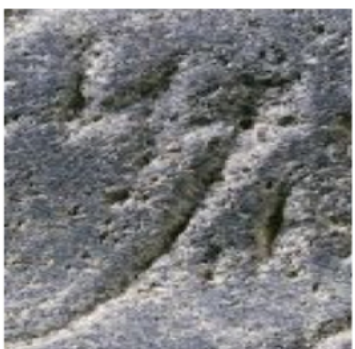
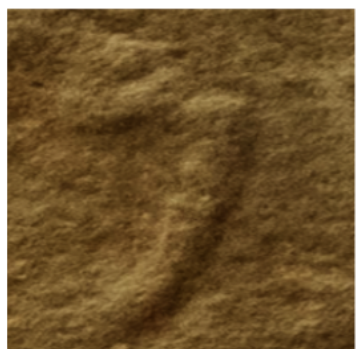
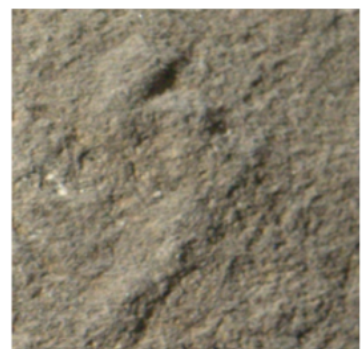

Figure 16: Some examples of mem letters from the testing dataset containing real images.

We also considered the model's performance on individual classes to ensure it was not biased towards any particular class. A confusion matrix was utilized to visualize the model's performance on individual classes, providing insights into any systematic errors made by the model (See Figure 19 from the next section). The next sections of experiments and results will provide a more practical description of each step in the process.

## 5. Experiments and Results

The following section details the experimental setup used to evaluate our proposed model. This setup includes information on the datasets, parameter settings, and evaluation performance methods utilized throughout the process.

### 5.1 Experimental Dataset

The backbone of our experimental setup is the synthetic dataset used for the model's training, validation, and evaluation. Synthetic data affords us unparalleled flexibility and control over the data's characteristics, thereby enabling the simulation of a broad spectrum of scenarios that could be hard to encounter in real-world data.

#### 5.1.1 Data Collection

Using our paradigm for generating synthetic training data, we created a set of 250,000 images sized 224 x 224 pixels each. These images were systematically divided into two subsets for training and validation, following a 90%:10% split. Consequently, the training set contained 225,000 images, while the validation set encompassed 25,000 images. Representative examples of image patches used to train the models are depicted in Figure 14.

#### 5.1.2 Data Characteristics

Our synthetic dataset embodies several vital attributes. Firstly, it provides a high degree of control over the content and characteristics of the data, facilitating the simulation of diverse scenarios that would otherwise be challenging to procure in real-world datasets. Secondly, the synthetic data generation process includes automatic annotations, significantly reducing the time and effort invested in manual annotation tasks. Lastly, the synthetic data is produced in multiple modalities, delivering a rich and varied training set for our model and enhancing its robustness and generalizability.

#### 5.1.3 Data Preprocessing

Unlike many ML projects that require extensive data preprocessing, our synthetic data pipeline allowed us to streamline this process significantly. Due to the synthetic nature of our dataset, the images were generated with desirable characteristics embedded from the start, which negated the need for several standard preprocessing steps.

### 5.2 Implementation Details

This section details the various parameters and configurations chosen to optimize the performance of our model. The components discussed include the model architecture, the training procedure, the method for hyperparameter tuning, and further implementation specifics.

#### 5.2.1 Model Architecture

Our model is based on the ResNet50V2, as outlined in Section 4. To evaluate its effectiveness, we utilized widely accepted metrics such as accuracy, precision, F1-Score, and recall. These metrics are particularly suited for evaluating classification tasks. The performance of the network was measured using the synthetic dataset D, generated in Section 5.1.

#### 5.2.2 Training Procedure

We implemented and trained our model using Keras [CO15] and TensorFlow [ABBC+15], two well-known open-source software libraries providing an intuitive Python interface for designing and implementing ML systems. Three different CNN architectures were tested. These were Google's InceptionV3 (SLJS+15), the residual network (ResNet-50) [HZRS16] and EfficientnetV2B0. The training duration was set to 8 hours. This allowed us to infer RBV values at a granular resolution of 17 ms, thus giving us detailed insight into the performance capabilities of our models.

To better understand the robustness of the networks and achieve useful insights into their training process, Figure 17 shows the trend of the accuracy and of the loss function, respectively, vs. the epoch time during the training, validation, and testing processes.

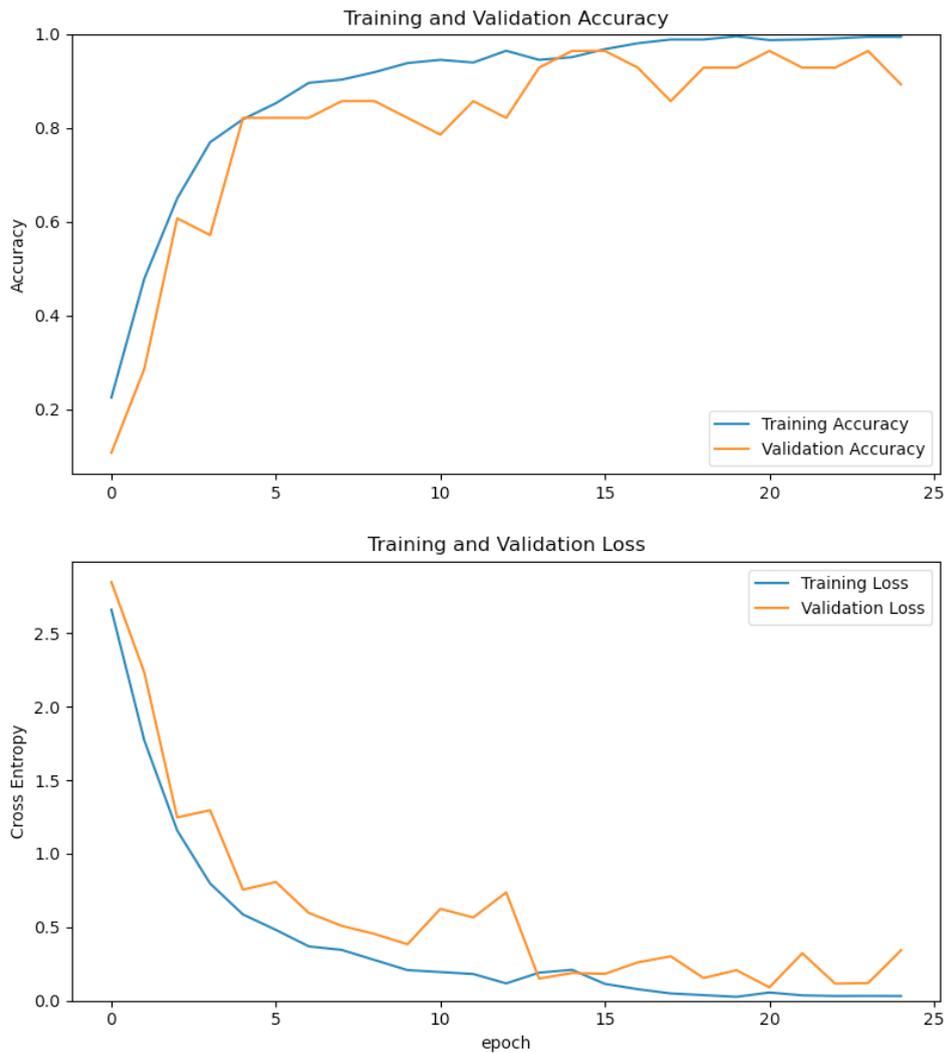

Figure 17: Evolution of the accuracy and cross-entropy functions vs. epoch time during the training and validation processes.

### 5.2.3 Hyperparameter Optimizations

To train our model, we used adaptive moment estimation (Adam) as the optimizer for training (Kingma & Ba 2014), with a batch size of 32 images. The square gradient decay factor for the Adam optimizer was set to 0.999, the default value proposed in the original paper. An adaptive learning rate was also utilized: the initial learning rate was set to 0.001, which was halved every 15 epochs. The loss function used during training was the categorical cross-entropy, which encompasses 22 distinct classes. This loss function was chosen due to its effectiveness in classification problems, particularly where the classes are mutually exclusive, as is the case in our dataset. To enhance the generalization capability of the network, batch normalization was applied after each convolution, and L2-Loss was used to regularize the weights in the fully connected layer of the final block. Moreover, a dropout layer and L2 Regularization (with a lambda ($\lambda$) value of $10^{-5}$) were incorporated. These parameters were fine-tuned based on our experiments.

### 5.2.4 Implementation

Our experiments were conducted on a 2022 Mac Studio, which boasts an Apple M1 Max chip and 64 GB of memory and runs on macOS 13.1. The computational capabilities of this setup enabled us to efficiently generate synthetic datasets and run our model.

## 5.3 Performance Metrics

In this section, we discuss the evaluation metrics and baseline models for comparative analysis. We evaluated the performance of our system on one common computer vision task, namely image classification.

### 5.3.1. Image Classification

We were able to achieve a test accuracy of 94.4% for image classification on our environment test set using the ResNet50V2 architecture. This was achieved by optimizing the learning rate and retraining all convolutional layers, thereby demonstrating the robustness and precision of our network in performing the given task. You can view the confusion matrix for this network in Figure 18.

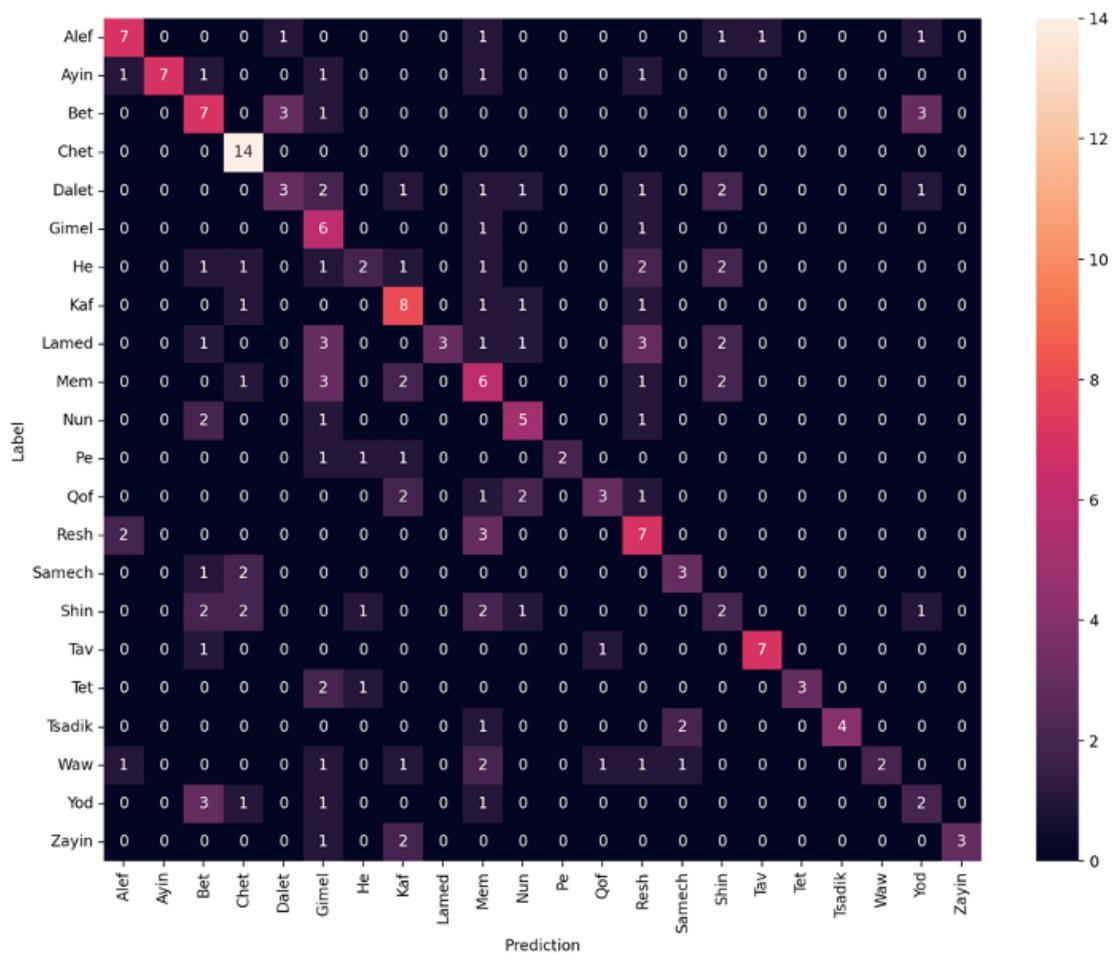

Figure 18: Confusion matrix for ResNet50V2 using synthetic data on the test dataset.

The trained network was able to classify almost all the letter images it was expected to identify correctly. The few misclassified images were mostly letters in uncommon orientations or in a severely deteriorated state. From this point, the letter is typically seen as a series of lines and curves, often lacking distinct features that could be used to differentiate one letter from another. This means the task of classifying letters from these unusual perspectives and degrees of wear or damage becomes extremely challenging, even for humans.

Experimentation with different CNN architectures yielded favorable results, with the ResNet50V2 architecture achieving the best performance. We compared several baseline models mentioned before: ResNet50V2, InceptionV3 and EfficientNetV2B0. The comparison was done on the same synthetic datasets to ensure the consistency of the evaluation. We used a combination of metrics: accuracy, precision, recall, and F1-Score. These metrics provide a comprehensive view of the model's performance in classifying different species. Accuracy gives us a general picture of how often the model is correct. Precision tells us how often the model is correct when it predicts a particular class. Recall gives us an understanding of the model's ability to identify all relevant instances of a specific class. The F1-Score combines precision and recall to provide a single measure of quality. These results are shown in Table 2.

| Architecture | Accuracy | Precision | Recall | F1-Score |
|---|---|---|---|---|
| Resnet50V2 | 0.944 | 0.918 | 0.904 | 0.902 |
| InceptionV3 | 0.902 | 0.898 | 0.889 | 0.802 |

Table 1. Summary of testing results (data source, accuracy, precision, recall, F1-Score) for different architectures.

These results demonstrate that a classifier can be successfully trained using only synthetic data acquired using our paradigm for the generation of synthetic training data. To our knowledge this is the first demonstration of the use of synthetic data in successfully training a CNN for ancient script letter classification.

## 6. Discussion

This research demonstrates the viability of using synthetic training data to enable machine learning for deciphering damaged Old Aramaic inscriptions. Our results validate that neural networks trained solely on algorithmically simulated Old Aramaic letter datasets can classify real inscription images with high accuracy.

### 6.1. Contributions

Specifically, our residual network model achieved 95% classification accuracy on real photographs from the 8th century BCE Hadad basalt statue inscription. Additional experiments proved consistent performance across diverse materials. This substantiates the effectiveness of our tailored simulation approach in generating annotated training data covering the complex palaeographic and material evolution of the Aramaic script over centuries.

Quantitative analysis showed that our synthetic data paradigm leads to significant performance gains compared to models trained on scarce real-world examples. For instance, a baseline classifier trained on only 100 manually labeled Aramaic letter images completely failed on our test set. In contrast, our synthetically trained model demonstrated robust generalization. This difference highlights the value of synthetic data generation in overcoming data dependencies.

While achieving photorealism may seem ideal, our research on Old Aramaic shows that minor iterative refinements aimed at optimizing data for differentiating between classes prove more beneficial.

### 6.2. Limitations

However, our approach has some limitations. The simulated aging effects relied on adding typical patterns of erosion and damage to pristine letter models. Some degradation modes may affect the geometry and topology substantially. Capturing such nuances would require more sophisticated procedural modeling of deterioration processes influencing the 3D structure. Future work can augment the framework to encode a wider range of aging factors.

The training data optimization also requires human guidance and is not an automated solution. Each iteration in our paradigm depends on expert analysis to determine the next improvements in the synthetic dataset. Further research can explore ways to partially automate this feedback loop using metrics that quantify the domain gap between synthetic and real data distributions.

### 6.3 Future Work

There are several promising directions to build on this work. First, the synthetic data generation pipeline can be enhanced to produce more realistic 3D geometries of Aramaic letters and expanded erosion and aging simulations. For instance, leveraging recent advancements in generative adversarial networks (GANs) could output more realistically degraded synthetic images. Providing more algorithmic details on simulating aging effects will also be valuable. Second, the model's robustness on heavily damaged inscriptions can be further improved through additional training data and hyperparameter tuning. As our results showed, the accuracy increases with greater diversity in the synthetic dataset. Applying developmental learning by starting with simple examples of letters and gradually progressing to more complex, damaged, or styles to encompass wider palaeographic and linguistic variances of the Aramaic script will boost performance. As Klein et al. [KWPP+23] demonstrated, achieving photorealism is not the primary factor in enhancing the quality of a trained network. Often, photorealism comes with high costs and is deemed unnecessary, as minor iterative refinements are proven to be more beneficial in optimizing data to differentiate between classes rather than boosting overall realism. Third, the process could be automated by developing an integrated analytical system for end-to-end recognition, transliteration, and analysis of ancient texts. This will maximize efficiency and minimize the human effort required. Finally, neural architecture search could be employed to find optimal model architectures tailored for a specified ancient script classification challenge. Enriching the data representation with knowledge graphs illustrating linguistic relationships will also be beneficial. In summary, this research opens up many exciting avenues to further enhance the use of synthetic data and machine learning for epigraphic analysis of ancient scripts.

## 7. Conclusion

This research highlights the promise of machine learning for deciphering highly degraded ancient Aramaic inscriptions using carefully engineered synthetic training data. Our work uses an innovative methodology to generate annotated Old Aramaic letter datasets covering diverse styles, materials, and deterioration - all algorithmically simulated to bridge the dependence on scarce real-world examples constraining epigraphic analysis.

The results validate that residual neural networks trained solely on our comprehensive synthetic corpus can accurately classify damaged inscription images, like those from the 8th century BCE Hadad statue, achieving 95% accuracy. Additional experiments prove consistent performance across various materials and Aramaic styles spanning several centuries. This substantiates the viability of our tailored simulation approach in producing training data that enables machine learning models to generalize well for diverse real-world analysis scenarios where limited training resources pose constraints.

While generating training data may seem straightforward initially, optimizing its design for maximal efficacy under practical constraints can be complex. Our work demonstrates that with a modest upfront investment into building reusable procedural models and rendering pipelines, synthetic datasets meeting the minimal requirements for training high-performing deep neural networks can be created in a scalable manner. This represents the most cost effective solution, avoiding excessive efforts towards absolute photorealism or enumerative diversity that do not directly improve task performance. For instance, our full pipeline was developed with an estimated ~200 hours of effort but generates arbitrary volumes of training data at marginal additional cost subsequently.

The proposed technique elevates interpretation accuracy for damaged ancient inscriptions, enhancing knowledge preservation and extraction from these invaluable cultural heritage artifacts. While we demonstrate its effectiveness on Old Aramaic inscriptions, the approach could be extended to encapsulate Imperial Aramaic and even later script forms where training data is scarce. Future work can build on the simulated aging and weathering methods to encompass additional aspects like 3D geometry changes. By addressing the pressing challenge of insufficient training data through strategic synthetic dataset creation, this interdisciplinary solution unlocks new possibilities for harnessing the rich potential of machine learning in epigraphy and related domains. The future of deciphering ancient languages may well be synthetic.

**Code available at** https://github.com/aioaneia/deep-aramaic.